\documentclass{article}

\usepackage{microtype}
\usepackage{graphicx}
\usepackage{subcaption}
\usepackage{booktabs} 
\usepackage{multirow}
\usepackage{tabularx}
\usepackage{xcolor}
\usepackage{colortbl}
\usepackage{arydshln}
\usepackage{pifont}
\usepackage{hyperref}

\newcommand{\itbf}[1]{\textit{\textbf{#1}}}

\newcommand{\tablerowcolor}{\rowcolor{gray!15}}

\newcommand{\eg}{\textit{e.g.}}
\newcommand{\ie}{\textit{i.e.}}

\usepackage[preprint]{icml2026}

\usepackage{amsmath}
\usepackage{amssymb}
\usepackage{mathtools}
\usepackage{amsthm}

\usepackage[capitalize,noabbrev]{cleveref}

\theoremstyle{plain}

\theoremstyle{definition}

\theoremstyle{remark}

\usepackage[textsize=tiny]{todonotes}

\icmltitlerunning{TranX-Adapter: Bridging Artifacts and Semantics within MLLMs for Robust AI-generated Image Detection}

\begin{document}

\twocolumn[
  \icmltitle{TranX-Adapter: Bridging Artifacts and Semantics within MLLMs for Robust\\ AI-generated Image Detection}



  \icmlsetsymbol{equal}{*}

  \begin{icmlauthorlist}
    \icmlauthor{Wenbin Wang}{whu,equal}
    \icmlauthor{Yuge Huang}{tencent}
    \icmlauthor{Jianqing Xu}{tencent}
    \icmlauthor{Yue Yu}{tencent} 
    \icmlauthor{Jiangtao Yan}{tencent} \\
    \icmlauthor{Shouhong Ding}{tencent} 
    \icmlauthor{Pan Zhou}{smu}
    \icmlauthor{Yong Luo}{whu}
  \end{icmlauthorlist}

  \icmlaffiliation{whu}{Wuhan University}
  \icmlaffiliation{tencent}{Tencent YouTu Lab}
  \icmlaffiliation{smu}{Singapore Management University}

  \icmlcorrespondingauthor{Pan Zhou and Yong Luo}{wangwenbin97@whu.edu.cn}


  \vskip 0.3in
]



\printAffiliationsAndNotice{*This work was done during an internship at Tencent YouTu Lab.}  

\begin{abstract}
  Rapid advances in AI-generated image (AIGI) technology enable highly realistic synthesis, threatening public information integrity and security.
  Recent studies have demonstrated that incorporating texture-level artifact features alongside semantic features into multimodal large language models (MLLMs) can enhance their AIGI detection capability. However, our preliminary analyses reveal that artifact features exhibit high intra-feature similarity, leading to an almost uniform attention map after the softmax operation. This phenomenon causes \texttt{attention dilution}, thereby hindering effective fusion between semantic and artifact features. To overcome this limitation, we propose a lightweight fusion adapter, \itbf{TranX-Adapter}, which integrates a \textit{Task-aware Optimal-Transport Fusion} that leverages the Jensen-Shannon divergence between artifact and semantic prediction probabilities as a cost matrix to transfer artifact information into semantic features, and an \textit{X-Fusion} that employs cross-attention to transfer semantic information into artifact features. Experiments on standard AIGI detection benchmarks upon several advanced MLLMs, show that our \itbf{TranX-Adapter} brings consistent and significant improvements (up to +6\% accuracy). Code will be available at \url{https://github.com/DreamMr/TranX-Adapter}.
\end{abstract}

\section{Introduction}
Recent progress in generative AI~\cite{karras2019style, dhariwal2021diffusion, goodfellow2014generative, rombach2022high, park2019semantic, zhu2017unpaired} has dramatically enhanced the capability to synthesize highly realistic images, enabling the creation of visual content that closely mimics real-world scenes. Despite these remarkable achievements, such technologies also pose substantial societal risks, as convincingly fake images can be exploited to mislead the public and propagate misinformation. These emerging threats have consequently driven the computer vision community to advance research in AI-generated image detection (AIGI detection)~\cite{tan2024rethinking, wang2020cnn, zhang2019detecting, qian2020thinking, liu2020global, ojha2023towards, yan2024sanity, chen2025dual}, aiming to develop more robust and reliable techniques for distinguishing synthetic image from authentic image.

\begin{figure}[t]
    \begin{center}
        \includegraphics[width=1.\linewidth]{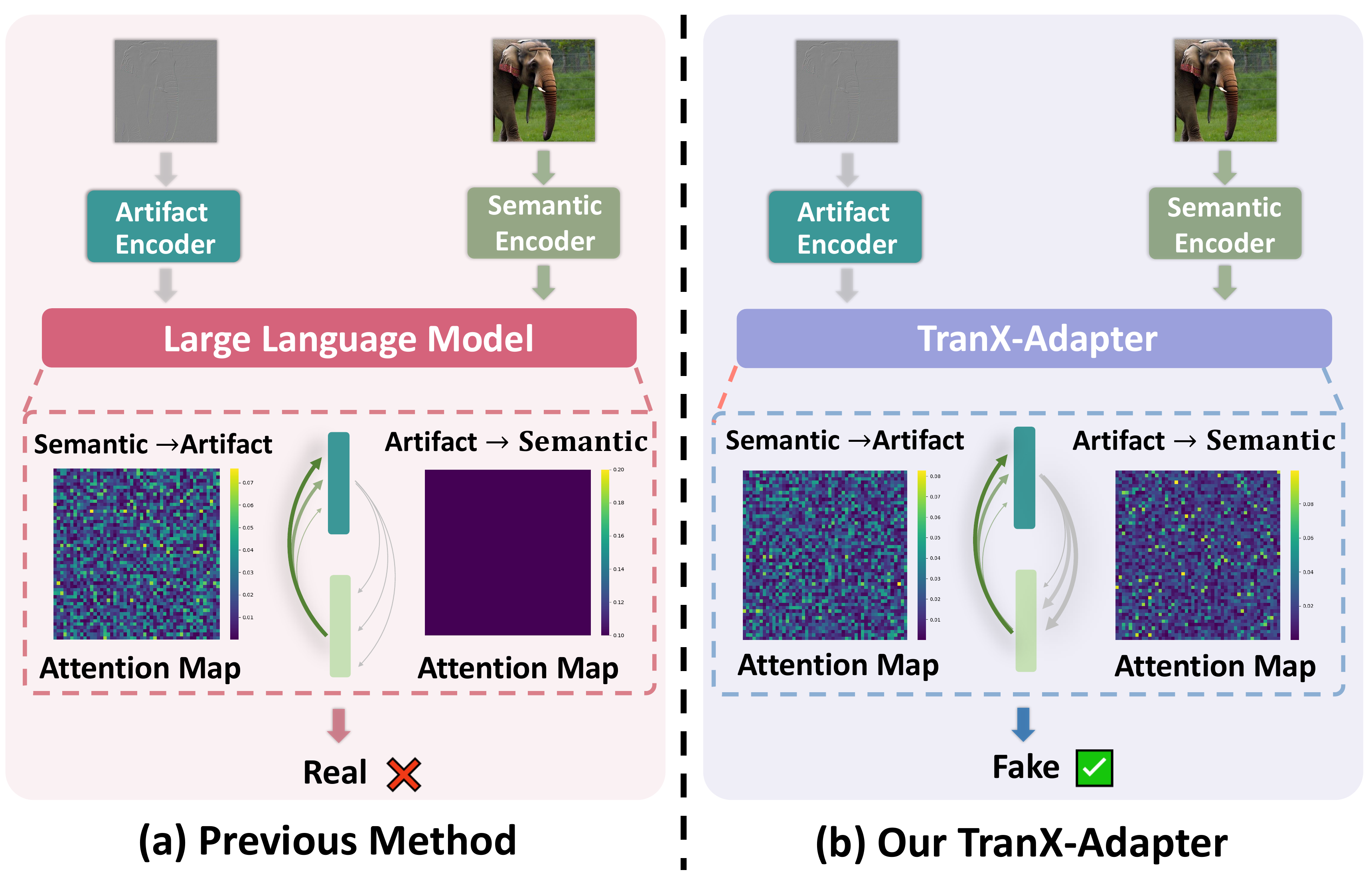}
    \end{center}
    \caption{\textbf{Comparison between the previous fusion method and our \itbf{TranX-Adapter}.} (a) Previous Method: Concatenates artifact (\eg, from NPR) and semantic features (\eg, from CLIP-ViT), resulting in uniform attention and weak interaction. (b) Our \itbf{TranX-Adapter}: Incorporates a lightweight bidirectional fusion mechanism that enhances feature interaction.}
    \label{fig:motivation}
    \vspace{-0.4cm}
\end{figure}

Despite substantial progress in AIGI detection, notable challenges persist. Early work~\cite{wang2020cnn, zhang2019detecting, tan2024rethinking} predominantly targets GAN and diffusion-generated images~\cite{goodfellow2014generative, dhariwal2021diffusion}, limiting robustness against the latest generative models~\cite{labs2025flux1kontextflowmatching}. To improve generalization, recent studies advance three complementary directions: (1) artifact-based methods~\cite{tan2024rethinking,qian2020thinking, zhong2023rich, yan2024sanity}, which capture pixel-level artifacts induced by up-sampling or interpolation; (2) semantic-based methods~\cite{tan2025c2p, yanorthogonal, xu2024fakeshield}, which leverage high-level visual semantics and broad world knowledge of Multimodal Large Language Models (MLLMs) to detect subtle or localized manipulations; and (3) hybrid artifact-semantic methods~\cite{zhou2025aigi, cheng2025co}, the most promising direction, explicitly combine artifact encoders with MLLMs to combine precise pixel-level cues with semantic robustness. A representative example is AIGI-Holmes~\cite{zhou2025aigi}, which integrates NPR~\cite{tan2024rethinking} with LLaVA-1.6-mistral~\cite{liu2024improved}, yielding state-of-the-art (SOTA) performance across diverse AIGI detection benchmarks.

However, our preliminary analyses reveal that naively concatenating artifact features (\eg, from NPR) with semantic features (\eg,  from CLIP-ViT) and feeding them into the Large Language Model (LLM) yields suboptimal fusion behavior. As illustrated in Figure~\ref{fig:motivation} (a), when artifact features are used as \textit{key} and \textit{value} and semantic features as \textit{query} to transfer artifact information into the semantic space (Artifact $\rightarrow$ Semantic), the resulting attention map collapses into an almost uniform pattern, manifesting the \texttt{attention dilution} phenomenon~\cite{zhang2024selective}. This degradation stems from the high intra-feature similarity of artifact representations, which suppresses discriminative cross-encoder interactions and limits the model's ability to effectively convey fine-grained, texture-level artifact cues.

To achieve more effective fusion between semantic and artifact features, we introduce a lightweight fusion adapter, termed \itbf{TranX-Adapter}, which is placed before the LLM and employs distinct fusion strategies for different interaction directions. The proposed \itbf{TranX-Adapter} comprises two  modules: \textit{Task-Aware Optimal-Transport Fusion (TOP-Fusion)} and \textit{X-Fusion}. Specifically, \textit{TOP-Fusion} transfers artifact features into the semantic features by optimal transport~\cite{cuturi2013sinkhorn}. We employ the distance between the task-specific prediction probabilities of the two features, instead of relying on the dot-product interaction used in self-attention~\cite{vaswani2017attention}. This design mitigates the \texttt{attention dilution} induced by the high intra-feature similarity among artifact features. In contrast, \textit{X-Fusion} transfer the semantic features into the artifact features by cross-attention. Here, we introduce \textit{X-Fusion} based on the observation that the interaction between the two features predominantly emerges in the shallow layers of the LLM. 
Accordingly, we confine the trainable parameters to a lightweight module, avoiding any modification to the LLM and improving training efficiency. 

Our contribution is identifying that fusing artifact and semantic features within MLLMs is hindered by the high intra-feature similarity of artifact representations, which weakens discriminative interactions. To address this challenge, we propose \itbf{TranX-Adapter}, a lightweight adapter that enables effective bidirectional fusion through \textit{TOP-Fusion} (Artifact $\rightarrow$ Semantic) and \textit{X-Fusion} (Semantic $\rightarrow$ Artifact). Extensive experiments show that \itbf{TranX-Adapter} consistently improves performance across both cross-method and cross-dataset benchmarks, achieving an average gain of 4.7\% and outperforming recent SOTA approaches.

\section{Related Work}
With the rapid progress of AI-based image generation, a broad spectrum of detectors has emerged. Recent studies advance three complementary directions: \textbf{(1) artifact-based methods}, \textbf{(2) semantic-based methods} and \textbf{(3) hybrid artifact-semantic methods.}

\noindent\textbf{Artifact-Based Methods.}
Artifact-Based methods primarily aim to capture texture-level features induced by the up-sampling operations inherent in image generation models~\cite{tan2024rethinking,zhong2023patchcraft,qian2020thinking,frank2020leveraging}. These features manifest as distinctive pixel-level artifacts, such as similar values between adjacent pixels.
\citet{tan2024rethinking} propose NPR to capture localized structural cues introduced by the up-sampling operations in CNN-based generative networks. 
~\citet{zhong2023patchcraft} introduce PatchCraft, a texture-patch-based method that suppresses global semantics and exploits inter-pixel correlation contrasts to improve generalization.

\noindent\textbf{Semantic-Based Methods.}
These methods identify synthetic images through semantic cues, such as human hand contours, by training deep neural networks on extensive real and generated datasets covering diverse object categories~\cite{xu2024fakeshield,chang2023antifakeprompt,tan2025c2p,chen2024drct}. A representative examples of these approaches are to leverage a powerful foundation model (\eg, MLLMs) that possesses extensive world knowledge, thereby endowing it with strong generalization capability. \citet{xu2024fakeshield} present FakeShield that leverages LLM to not only detect image forgery and localize tampered regions but also produce human-readable rationales for its judgments. \citet{chang2023antifakeprompt} formulate deepfake image detection as a visual question-answering task and leverages prompt-tuned MLLM to significantly improve generalisability to unseen generative models.

\noindent\textbf{Hybrid Artifact-Semantic Methods.}
These methods simultaneously integrate the previous two approaches by combining pixel-level artifact cues with semantic world knowledge, achieving notable performance gains~\cite{cheng2025co,zhou2025aigi}. \citet{cheng2025co} introduce a hybrid method CO-SPY that jointly enhances and adaptively fuses semantic and pixel-artifact features to robustly detect synthetic images across diverse generative models. AIGI-Holmes~\cite{zhou2025aigi} fuses NPR with semantic-rich pretrained CLIP-ViT embedding within MLLM to deliver both robust generalisation to unseen AI-generated images and human verifiable explanations.

However, our preliminary analysis (Sect.~\ref{sec:pilot_study}) finds that directly concatenating artifact features and semantic features as input to the LLM leads to suboptimal fusion, since the artifact features exhibit high intra-feature similarity, resulting in an almost uniform attention map that weakens interaction. To address this limitation, we introduce \itbf{TranX-Adapter} in Sect.~\ref{sec:method}, which promotes a more profound integration of artifact and semantic representations, thereby advancing the effectiveness and robustness of AIGI detection.

\section{Pilot Study}
\label{sec:pilot_study}

In this section, we conduct a pilot study to demonstrate that self-attention fusion of artifact and semantic features within the LLM is suboptimal. Following AIGI-Holmes~\cite{zhou2025aigi}, we employ NPR~\cite{tan2024rethinking} as the artifact encoder and integrate it into LLaVA-1.6-mistral 7B~\cite{liu2024improved}, while CLIP-ViT provides the semantic features. The two types of features are concatenated and fed into the LLM for joint training. In Sect.~\ref{sec:feature}, we analyze the intrinsic distributional differences between artifact and semantic features. In Sect.~\ref{sec:attention}, attention map visualization reveals that when the LLM performs self-attention fusion, the Artifact $\rightarrow$ Semantic interaction exhibits \texttt{attention dilution}~\cite{zhang2024selective}, where the attention distribution becomes nearly uniform, indicating ineffective feature fusion. Finally, by quantifying the mean significance of information flow, we further confirm that existing self-attention fusion within the LLM remains suboptimal (Sect.~\ref{sec:attention}).

\begin{figure}[hbt]
    \begin{center}
        \includegraphics[width=1.\linewidth]{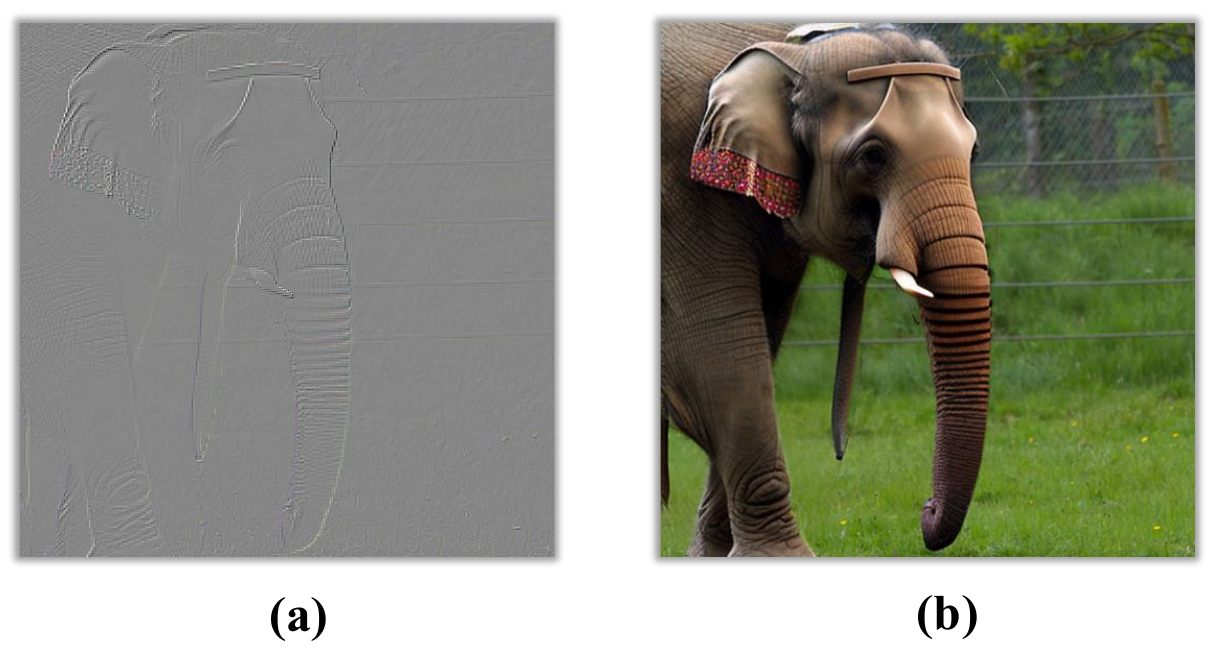}
    \end{center}
    \caption{\textbf{Comparison between the processed NPR input (a) and the CLIP-ViT input image\protect\footnotemark~ (b)}, where NPR highlights local pixel interdependencies for synthetic image detection.}
    \label{fig:input_compare}
\end{figure}

\footnotetext{The image is sourced from BiasFree~\cite{guillaro2025bias}}

\subsection{Uniform Artifact and Variance Semantic}
\label{sec:feature}

In AIGI detection, artifact features are essential for distinguishing synthetic images from authentic ones~\cite{tan2024rethinking, chen2025dual, qian2020thinking}. The NPR model exemplifies this principle by leveraging local pixel dependencies to capture subtle inconsistencies that signal synthetic content. Specifically, NPR first upsamples the input image through bilinear interpolation, subsequently downsamples it, and computes the residual between the reconstructed and original images, which is then processed by ResNet~\cite{he2016deep}.

\begin{figure}[hbt]
    \begin{center}
        \includegraphics[width=1.\linewidth]{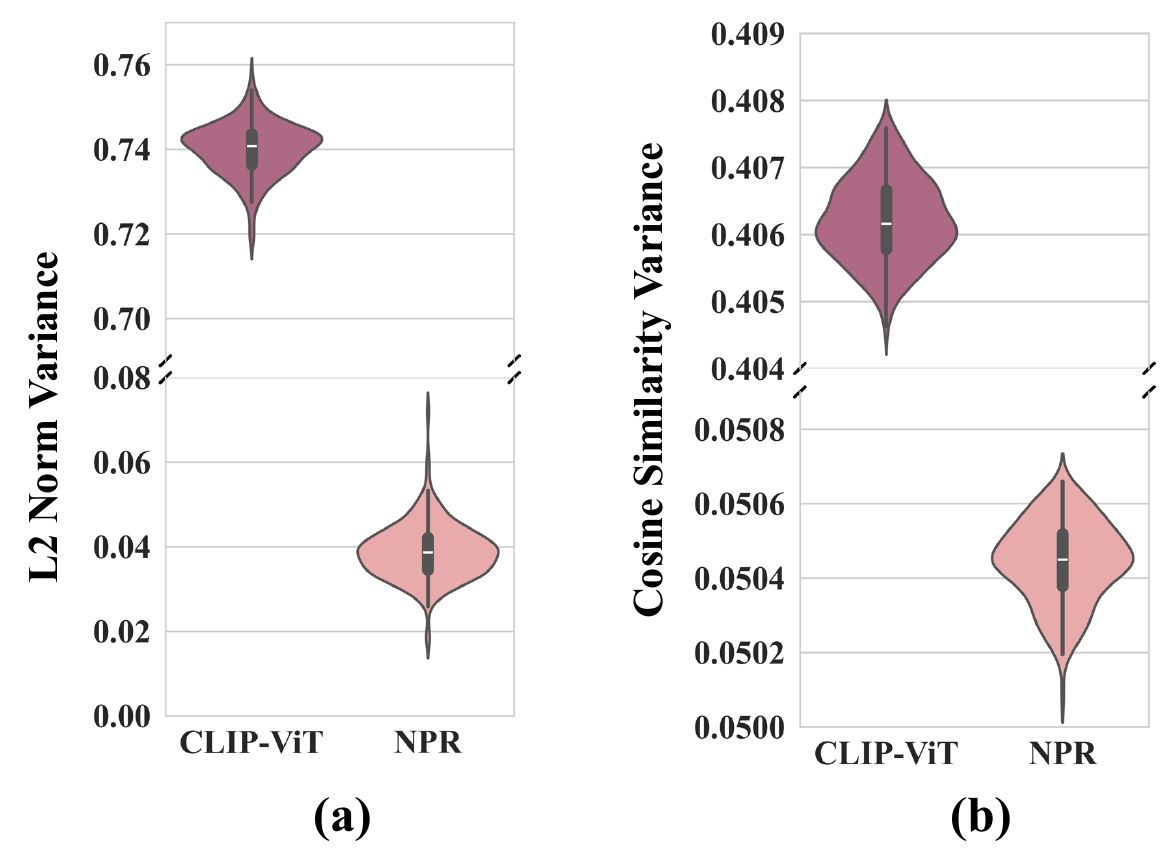}
    \end{center}
    \caption{\textbf{Distributional comparison of representational variances between CLIP-ViT and NPR:} (a) variance of L2 norms and (b) variance of cosine similarities across image patches.}
    \label{fig:fea_variance}
\end{figure}

As illustrated in Figure~\ref{fig:input_compare} (a), the NPR input image primarily captures structural cues such as edges and textures, leading to substantial redundancy among spatial patches. In contrast, MLLMs typically employ CLIP-ViT as the visual encoder, which directly processes the original image to extract rich and diverse semantic representations. This inherent discrepancy is further manifested at the feature level. As shown in Figure~\ref{fig:fea_variance}, we examine the distributional characteristics of artifact and semantic features by computing the L2 norm and cosine similarity image patches derived from each visual encoder. The results reveal that CLIP-ViT features exhibit significantly higher inter-patch variance, whereas NPR features demonstrate strong intra-feature similarity, indicating that \itbf{artifact representations are more homogeneous and less discriminative in feature space.}

\begin{figure*}[thbp]
    \begin{center}
        \includegraphics[width=1.\linewidth]{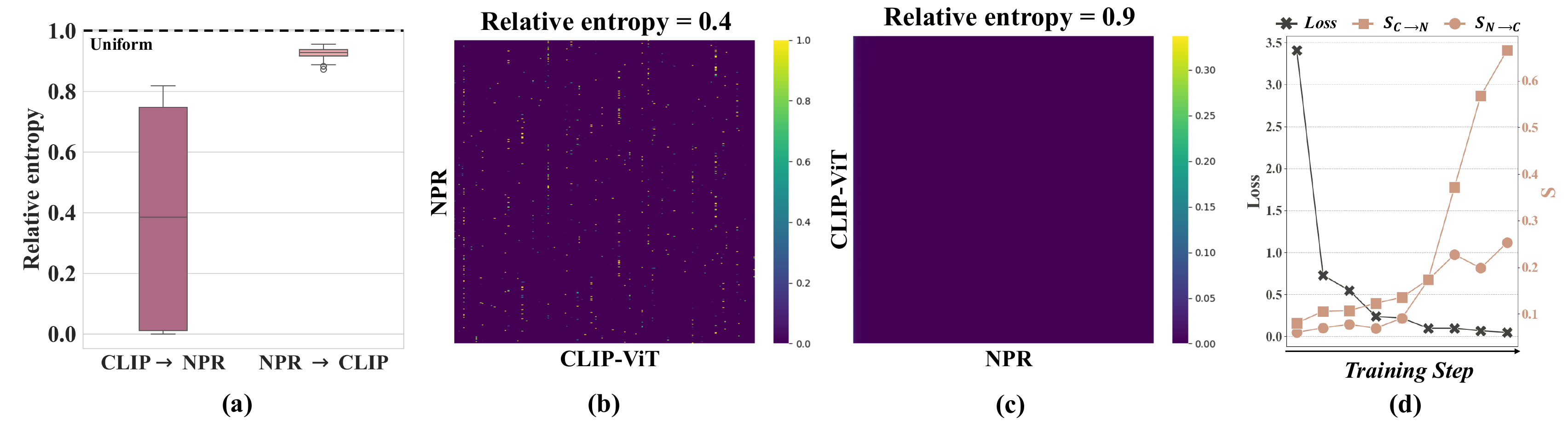}
    \end{center}
    \caption{\textbf{Comparison of cross-encoder interactions between CLIP-ViT and NPR.} (a) Relative entropy of attention maps. Higher relative entropy values, approaching $1$, indicate that the distribution is closer to uniform. (b) Attention map for the direction $CLIP\rightarrow NPR$, where the \texttt{query} originates from NPR features and the \texttt{key} and \texttt{value} are derived from CLIP-ViT features. (c) Attention map for the direction $NPR\rightarrow CLIP$, where the \texttt{query} corresponds to CLIP-ViT features and the \texttt{key} and \texttt{value} are obtained from NPR features. (d) Relationship between the training loss and the information flow metrics $S$.}
    \label{fig:distribution}
\end{figure*}

\subsection{Semantic Limitations in Artifact Cue Extraction}
\label{sec:attention}

We previously observed that the artifact features extracted by NPR exhibit strong intra-feature similarity. Consequently, during self-attention, this high similarity causes the attention map, when semantic features are used as \textit{queries} and artifact features as \textit{keys}, to collapses into an almost uniform distribution after softmax normalization. As illustrated in Figure~\ref{fig:distribution} (a), the relative entropy of the CLIP-ViT $\rightarrow$ NPR is notably lower, indicating a more concentrated distribution as shown in Figure~\ref{fig:distribution} (b). In contrast, the NPR $\rightarrow$ CLIP-ViT exhibits higher relative entropy, resulting in an almost uniform pattern as shown in Figure~\ref{fig:distribution} (c) and leading to the phenomenon of \texttt{attention dilution}. However, the critical forgery cues embedded in artifact features typically reside in high-frequency regions~\cite{zhong2023patchcraft}, and this diluted attention impedes the effective transmission of critical forgery cues from artifact to semantic.

To further substantiate this observation, we follow~\citet{wang2023label} and introduce two quantitative metrics: (1) $S_{N\rightarrow C}$ denotes the mean significance of the information flow from NPR to CLIP-ViT, and (2) $S_{C\rightarrow N}$ denotes the mean significance of information flow from CLIP-ViT to NPR. As shown in Figure~\ref{fig:distribution} (d), the information flow $S_{N\rightarrow C}$ is substantially lower than $S_{C\rightarrow N}$, indicating that \itbf{the artifact information encoded by NPR is difficult to transfer into the semantic feature space}.

To confirm that this phenomenon is not specific to NPR, we further observe similar behavior in VAE-based artifact encoder~\cite{cheng2025co}. These findings highlight that effectively bridging artifact and semantic features remains a critical challenge for MLLM performance in AIGI detection tasks.

\begin{figure*}[t]
    \begin{center}
        \includegraphics[width=1.\linewidth]{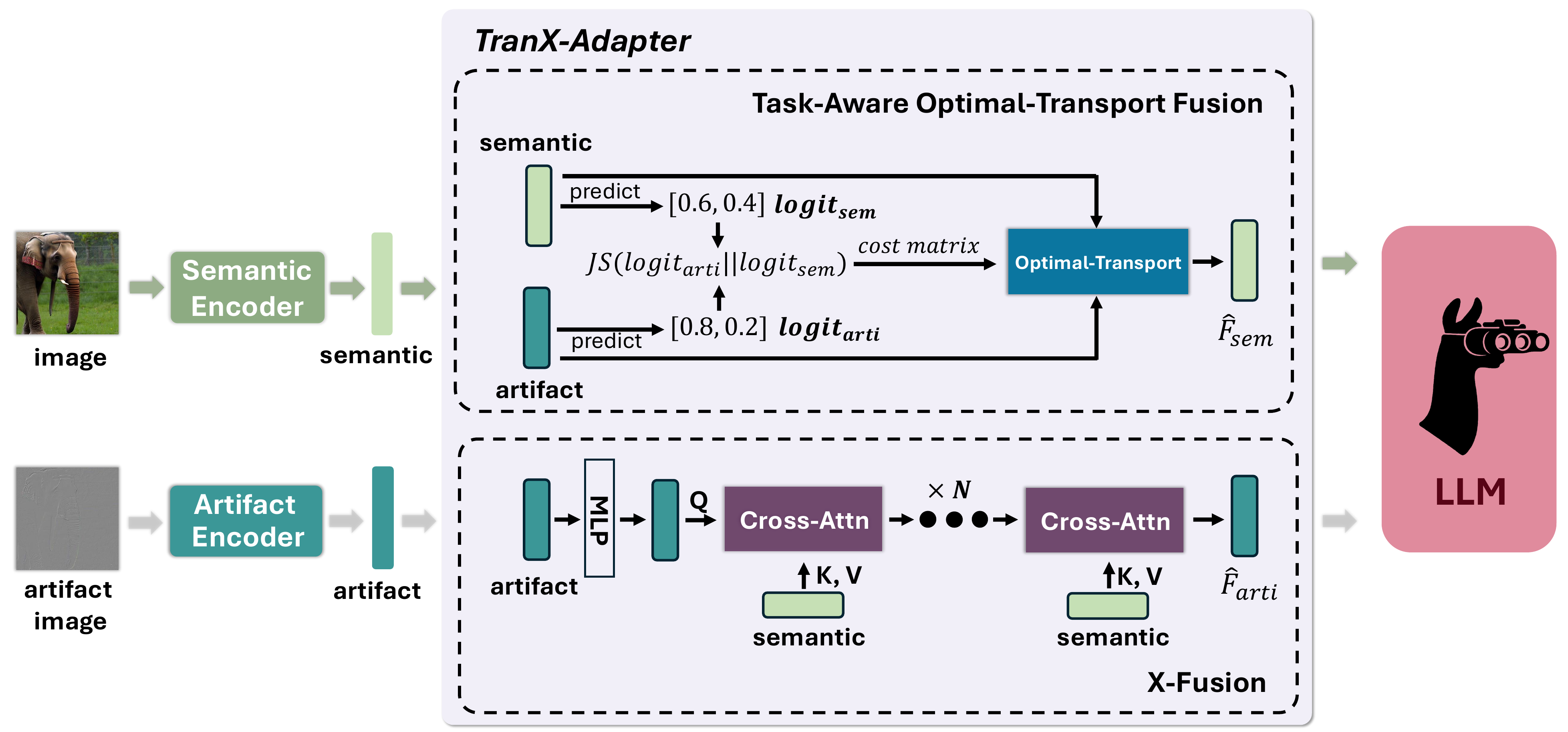}
    \end{center}
    \vspace{-0.2cm}
    \caption{\textbf{Overview of the proposed \itbf{TranX-Adapter}.} Our \itbf{TranX-Adapter} consists of two complementary fusion modules. The \textit{Task-Aware Optimal-Transport Fusion (TOP-Fusion)} aligns artifact and semantic feature predictions by computing JS divergence between their logits and transferring artifact features into the semantic features through optimal-transport, yielding an enhanced semantic feature $\hat{F}_{sem}$. The \textit{X-Fusion} module transfers semantic features into artifact features via multi-layer cross-attention, producing $\hat{F}_{art}$. The fused representations are finally fed into the Large Language Model (LLM) for detection.}
    \label{fig:method}
\end{figure*}

\section{Method}
\label{sec:method}
With the insights in Sect.~\ref{sec:pilot_study}, we introduce a lightweight fusion adapter, termed \itbf{TranX-Adapter}, which facilitates more effective fusion of semantic and artifact features. In this section, we introduce the \itbf{TranX-Adapter} in detail.

\subsection{Preliminary}
\label{sec:preliminary}

Given an image, we first extract artifact features $F_{art} \in \mathbb{R}^{N\times D}$ and semantic features $F_{sem} \in \mathbb{R}^{M\times D}$ from artifact encoder (\eg, NPR) and semantic encoder (\eg, CLIP-ViT) respectively. The $N$ and $M$ represent the lengths of the sequences of visual patches for artifact features $F_{art}$ and semantic features $F_{sem}$. The $D$ is the feature dimension of each visual patches. In fact, we employ separate Multi-Layer Perceptron (MLP) to align the feature dimensions of the artifact features and semantic features. And then the artifact features $F_{art}$ and semantic features $F_{sem}$ are fused by our \itbf{TranX-Adapter}, which enables bidirectional fusion via \textit{Task-Aware Optimal-Transport Fusion} (Artifact $\rightarrow$ Semantic) and \textit{X-Fusion} (Semantic $\rightarrow$ Artifact). The fused visual features are projected to the LLM as visual tokens $\mathcal{V}$ and paired with a textual prompt $T$. To train our \itbf{TranX-Adapter}, we apply the language modeling objective $\mathcal{L}$ to predictions of text tokens:
\begin{equation}
    \mathcal{L}=-\sum_{t} \log P_\theta(y_t|\mathcal{V}, T, y_{<t}).
\end{equation}

\subsection{Task-Aware Optimal-Transport Fusion}
\label{sec:top_fusion}

In this part, we first provide the motivation of our \textit{Task-Aware Optimal-Transport Fusion (TOP-Fusion)}, and then introduce it in detail.

\vspace{0.3cm}

\noindent\textbf{Motivation of TOP-Fusion.}

The primary objective of this module is to transfer discriminative artifact cues into semantic features, compensating for forgery cues that the semantic space fails to capture.
In Sect.~\ref{sec:pilot_study}, we observe that the Artifact $\rightarrow$ Semantic interaction suffers from \texttt{attention dilution}, where the attention distribution becomes nearly uniform due to the high intra-feature similarity of artifact features. Prior studies~\cite{zhong2023patchcraft, frank2020leveraging} indicate that generative models predominantly introduce artifacts in the high-frequency regions of synthetic images, leading to varying artifact intensity across patches. To capture this, we replace the artifact features with the probability of each patch being predicted as fake, and similarly map semantic features into the same probability space. The fusion process is guided to emphasize regions exhibiting substantial discrepancies between the artifact and semantic features. 

\vspace{0.2cm}

\noindent\textbf{TOP-Fusion.}
The processes of our \textit{TOP-Fusion} is illustrated in Figure~\ref{fig:method}. As discussed above, we first convert both artifact and semantic features into the probability of each patch being predicted as ``fake''. For artifact encoder, we use the prediction head from the artifact encoder training to obtain logits over the detect ``fake''. For the semantic encoder, when CLIP-ViT is used, we encode the text prompt ``\textit{A photo is a fake.}'' with CLIP's text encoder~\cite{devlin2019bert} and compute its similarity to the CLIP-ViT visual features. For semantic encoders without an aligned text encoder, we use a pretrained prediction head to obtain logits for fake detection. The complementary ``real'' logit is obtained as $1-\sigma(\cdot)$. The $\sigma(\cdot)$ is the sigmoid function. We denote $P$ and $S$ as the probability distributions corresponding to the artifact and semantic features, respectively. Then, we quantify the discrepancy between artifact and semantic using the Jensen-Shannon (JS) divergence, \ie, $JS(S||P)$.

Next, we transfer the artifact features into the semantic space based on this discrepancy. Specifically, we employ Optimal-Transport~\cite{peyre2019computational, santambrogio2015optimal}, using the $JS(S||P)$ as the cost matrix, and obtain the transport plan $\gamma$ via the Sinkhorn algorithm~\cite{cuturi2013sinkhorn}. Since the patches with substantial discrepancies should receive greater emphasis during feature transfer, we adopt the negative JS divergence, \ie, $-JS(S||P)$. Finally, we use the transport plan $\gamma$ to transfer the artifact features into semantic features:
\begin{gather}
    F_{art\rightarrow sem}=\gamma (F_{art}W_{art}), \\
    \hat{F}_{sem} = F_{sem} + \texttt{MLP}(F_{art\rightarrow sem}),
\end{gather}
where $W_{art} \in \mathbb{R}^{D\times d}$ are learned linear projections into the shared space, and $\texttt{MLP}(\cdot)$ is the Multi-Layer Perceptron that adapts the $F_{art\rightarrow sem}$ before fusion.

\subsection{X-Fusion}
\label{sec:x_fusion}

In this part, we first provide the motivation of our \textit{X-Fusion}, and then introduce it in detail.

\vspace{0.3cm}

\noindent\textbf{Motivation of X-Fusion.}
Previously, we introduced \textit{TOP-Fusion} to achieve the interaction from Artifact $\rightarrow$ Semantic. The most straightforward approach is to directly integrate \textit{TOP-Fusion} into the LLM. However, we argue that such a strategy is inefficient for two main reasons: (1) modifying the internal architecture of the LLM may disrupt its inherent knowledge, and (2) it requires training the  LLM. In fact, our experiments (Sect.~\ref{sec:discussion}) reveal that the interactions among different visual features within the LLM primarily occur in the shallow layers. Therefore, we concentrate the interaction between artifact and semantic features within a lightweight adapter and fine-tune only this adapter during training, avoiding the need to update the LLM. Consequently, we employ a cross (X)-attention mechanism to achieve the Semantic $\rightarrow$ Artifact interaction.

\vspace{0.2cm}

\noindent\textbf{X-Fusion.}
Given the artifact features $F_{art}$ and semantic features $F_{sem}$, both feature types are first projected into a shared latent space of dimension $d$ through linear transformations:
\begin{equation}
    \tilde{F}_{art}=F_{art}\tilde{W}_{art}, \ \quad\tilde{F}_{sem}=F_{sem}\tilde{W}_{sem},
\end{equation}
where $\tilde{W}_{art}, \text{ and } \tilde{W}_{sem} \in \mathbb{R}^{D\times d}$. Subsequently, a stack of cross-attention layers is employed to enable fine-grained semantic injection. In each layer $\ell$, the artifact features serve as the \textit{query}, while semantic features act as \textit{key} and \textit{value}:
\begin{gather}
    Q=\tilde{F}_{art}W_Q,\  K=\tilde{F}_{sem}W_K,\  V=\tilde{F}_{sem}W_V,\\
    H=\texttt{softmax}(\frac{QK^\top}{\sqrt{d}})V,\\
    X = \tilde{F}_{art} + H, \\
    F_{sem\rightarrow art} = X + \texttt{MLP}(X),
\end{gather}
where $W_Q, W_K \text{~and~} W_V \in \mathbb{R}^{d\times d}$ are the parameter matrics. In this formulation, the artifact features actively retrieve complementary semantic cues from the semantic features, facilitating semantic-aware enhancement. After the final cross-attention block, the updated representation is normalized and mapped back to the original feature dimension:
\begin{equation}
    \hat{F}_{art} = F_{art} + \texttt{MLP}(F_{sem\rightarrow art}).
\end{equation}
This process yields a refined artifact features that effectively incorporates semantic guidance while preserving the original feature characteristics.

\section{Experiments}

In this section, we first present the experimental setup, including implementation details and evaluation protocols, followed by the experiments and analysis. 

\subsection{Experimental Setup}

\paragraph{Implementation Details.} 
Following~\cite{zhou2025aigi}, we adopt NPR~\cite{tan2024rethinking} as the artifact encoder to extract artifact features, while using the visual encoder of the MLLM as the semantic encoder to capture high-level visual semantics. We use LLaVA-1.6-mistral 7B and Qwen3-VL (2B and 4B) as the base models. Unless otherwise specified, the parameter of MLLMs are kept frozen during training. 

\paragraph{Evaluation Protocols and Dataset.}
We adopt two widely recognized evaluation protocols. In \textbf{Protocol I}, we follow the GenImage~\cite{zhu2023genimage} benchmark and train the model on ImageNet~\cite{deng2009imagenet} real images and SD v1.4~\cite{rombach2022high} generate images, and evaluate the trained model across eight different generators~\cite{brock2018large,nichol2021glide, gu2022vector,dhariwal2021diffusion,midjourney, wukong,rombach2022high}. In \textbf{Protocol II}, the model is trained on data from diverse generative models and evaluated on comprehensive benchmarks containing challenging samples from modern generators. This protocol uses the Chameleon~\cite{yan2024sanity} and RRDataset~\cite{li2025bridging}.

\subsection{Comparison with AIGI Detection Methods}
\noindent\textbf{Comparison on GenImage (Protocol I).}

As shown in Table~\ref{tab:GenImage}, \itbf{TranX-Adapter} consistently boosts performance across all evaluated models, including LLaVA-1.6-mistral 7B, Qwen3-VL 2B, and Qwen3-VL 4B, demonstrating strong model-agnostic effectiveness.
Additionally, many existing methods overfit to specific generators and suffer significant performance drops on unseen ones, whereas our method maintains consistently high accuracy across generators, demonstrating strong robustness and transferability.

\begin{table*}[]
\caption{\textbf{Cross-model accuracy performance on the GenImage Dataset.} Accuracy (\%) of different detectors (rows) in distinguishing real images from those produced by various generative models (columns). The best result are marked in \textbf{bold}. ``$\dagger$'' indicates direct concatenation of artifact and semantic features following~\citet{zhou2025aigi}.}

\centering
\resizebox{1.\linewidth}{!}{
\begin{tabular}{lccccccccc}
\toprule
\itbf{Method}      & \itbf{Midjourney} & \itbf{SD v1.4} & \itbf{SD v1.5} & \itbf{ADM}  & \itbf{GLIDE} & \itbf{Wukong} & \itbf{VQDM} & \itbf{BigGAN} & \itbf{Mean} \\ \hline
F3Net~\cite{qian2020thinking}       & 50.1       & \textbf{99.9}    & \textbf{99.9}    & 49.9 & 50.0  & \textbf{99.9}   & 49.9 & 49.9   & 68.7 \\

DIRE~\cite{wang2023dire} & 60.2 & 99.9 & 99.8 & 50.9 & 55.0 & 99.2 & 50.1 & 50.2 & 70.7 \\
PatchCraft~\cite{zhong2023rich}  & 79.0       & 89.5    & 89.3    & 77.3 & 78.4  & 89.3   & 83.7 & 72.4   & 82.3 \\
UnivFD~\cite{ojha2023towards} & 73.2 & 84.2 & 84.0 & 55.2 & 76.9 & 75.6 & 56.9 & 80.3 & 73.3 \\
AIDE~\cite{yan2024sanity}        & 79.4       & 99.7    & 99.8    & 78.5 & \textbf{91.8}  & 98.7   & 80.3 & 66.9   & 86.9 \\
NPR~\cite{tan2024rethinking}         & 89.8       & 90.7    & 90.7    & 84.6 & 90.3  & 90.7   & 87.0 & 81.8   & 88.3 \\
LLaVA-1.6-mistral 7B$^\dagger$~\cite{liu2024improved} &   88.6         &  94.0       &   94.0      &  80.3   &   86.8   &   93.4   &  82.6   &   76.2    &  87.3  \\
AIGI-Holmes~\cite{zhou2025aigi} &     81.6       &     91.3    &    91.4     &   \textbf{88.4}   &   91.5    &   89.5     &   \textbf{90.9}   &     \textbf{94.5}   &   89.8   \\
Qwen3-VL 2B$^\dagger$~\cite{Qwen3-VL} &     88.1       &   87.3      &   84.1      &  78.6   &   81.7   &   81.7   &  84.1   &   72.2    &  82.2  \\
Qwen3-VL 4B$^\dagger$~\cite{Qwen3-VL} &     87.3       &    98.4     &    96.0     &  65.1   &   91.3   &   96.0   &  73.8   &   78.6    &  85.8  \\ \hline
\multicolumn{10}{c}{\itbf{w/ our TranX-Adapter}} \\ \hline

Qwen3-VL 2B &    90.5        &     97.6    &    96.0     &   83.3   &   89.7    &    92.9    &   82.5   &     71.4   &  88.0   \\
Qwen3-VL 4B &    92.1        &   97.6     &     97.6   &    81.0  &  83.3     &   94.4     &  84.9    &   87.3     &  89.8   \\
\tablerowcolor \textbf{LLaVA-1.6-mistral 7B} &     \textbf{94.9}       &    96.4     &    96.4     &    87.0  &   88.0    &   94.9     &   90.1   &     85.9   &  \textbf{91.9}   \\ 
\bottomrule
\end{tabular}
}
\label{tab:GenImage}
\end{table*}

\noindent\textbf{Comparison on Chameleon (Protocol II).}
To mitigate potential biases arising from training configurations, such as generator diversity and image category imbalance, we evaluate our method alongside existing detectors under varied training conditions. As shown in Table~\ref{tab:Chameleon}, prior approaches exhibit limited transferability, with accuracies ranging from 55\% to 72\%. In contrast, LLaVA-1.6-mistral 7B w/ our \itbf{TranX-Adapter} achieves 75.8\% accuracy when trained solely on SD v1.4 and 85.1\% when trained on the full GenImage dataset, significantly outperforming all baselines. These results underscore the strong generalization and robustness of the proposed \itbf{TranX-Adapter}.

\begin{table}[htb]
\caption{\textbf{Cross-dataset accuracy performance on the Chameleon testset.} ``$\dagger$'' denotes the setting following~\citet{zhou2025aigi}, where the artifact features and semantic features are directly concatenated. 
}
\centering
\resizebox{1.\linewidth}{!}{
\begin{tabular}{lcc}
\toprule
  \multirow{2}{*}{\itbf{Method}}            & \multicolumn{2}{c}{\itbf{Training Set}}              \\ \cmidrule(lr){2-3}
              & \itbf{SDv1.4}       & \itbf{All GenImage} \\ \hline
UnivFD~\cite{ojha2023towards}        &     55.6         & 60.4         \\
DIRE~\cite{wang2023dire}          &       59.7       & 57.8         \\
PatchCraft~\cite{zhong2023rich}    &        56.3      & 55.7         \\
NPR~\cite{tan2024rethinking}           &        58.1      & 57.8         \\
LLaVA-1.6-mistral 7B$^\dagger$~\cite{liu2024improved} & 69.4  &  81.9 \\
AIDE~\cite{yan2024sanity}          &        62.6      & 65.8         \\
PatchAll/CLIP~\cite{yang2025all} &         63.9     & 69.3         \\ 
PatchAll/DINOv2~\cite{yang2025all} &         66.6     & 72.1         \\
Qwen3-VL 2B$^\dagger$~\cite{Qwen3-VL}  & 68.5  & 78.8  \\
Qwen3-VL 4B$^\dagger$~\cite{Qwen3-VL}  & 69.4  &  78.3 \\ \hline
\multicolumn{3}{c}{\itbf{w/ our TranX-Adapter}} \\ \hline
Qwen3-VL 2B &     71.8         &    82.3       \\
Qwen3-VL 4B &     72.6         & 83.6        \\
\tablerowcolor \textbf{LLaVA-1.6-mistral 7B} &     \textbf{75.8}         & \textbf{85.1}        \\
\bottomrule
\end{tabular}
}
\label{tab:Chameleon}
\vspace{-0.2cm}
\end{table}

\noindent\textbf{Comparison on RRDataset (Protocol II).}
Compared with Chameleon, RRDataset~\cite{li2025bridging} encompasses seven distinct scenes and four re-digitization processes, offering a more comprehensive assessment of the generalization capability of AIGI detectors. As presented in Table~\ref{table:RRDataset}, we evaluate both zero-shot MLLMs and conventional detectors on RRDataset. Our \itbf{TranX-Adapter} integrated with Qwen3-VL 4B achieves the highest accuracy of 90.9\%, surpassing the strongest baseline (Qwen3-VL 4B, 85.6\%) by 5.3\% and GPT-4o by +6.8\%. These results demonstrate the superior generalization of our approach for AIGI detection.

\begin{table}[thb]
\caption{\textbf{Cross-dataset accuracy performance on the RRDataset.} The ``\itbf{Ori.}'' represents the Original, ``\itbf{Trans.}'' represents the Transmission, and ``\itbf{Re.}'' represents the Re-digitization. ``$\dagger$'' indicates the direct concatenation of artifact and semantic features~\cite{zhou2025aigi}.}
\centering
\resizebox{1.\linewidth}{!}{
\begin{tabular}{lcccc}
\toprule
\multirow{2}{*}{\itbf{Method}}              & \multicolumn{4}{c}{\itbf{RRDataset}}                               \\
       \cmidrule(lr){2-5}              & \itbf{Ori.}   & \itbf{Trans.}   & \itbf{Re.}   & \itbf{Avg.}   \\ \hline
\multicolumn{5}{c}{\textit{MLLMs (Zero-shot)}}                                               \\ \hline
GPT-4o~\cite{achiam2023gpt}               & 94.5       & 84.7           & 73.1              & 84.1      \\
Claude-3.7-sonnet~\cite{Anthropic2024}    & 89.9       & 83.8           & 73.9              & 82.5      \\
Gemini-2-flash~\cite{team2023gemini}       & 85.3       & 74.8           & 71.8              & 77.3      \\
Qwen2VL-72B~\cite{yang2024qwen2}          & 59.9       & 56.4           & 59.8              & 58.7    \\ \hline
\multicolumn{5}{c}{\textit{Detectors (Fine-tuned)}} \\ \hline
DIRE~\cite{wang2023dire}                 & 94.0       & 94.1           & 50.2              & 79.4      \\
AIDE~\cite{yan2024sanity}                 & 79.0       & 76.8           & 79.6              & 78.4      \\
NPR~\cite{tan2024rethinking}                  & 72.7       & 62.6           & 65.6              & 67.0      \\
LLaVA-1.6-mistral 7B$^\dagger$~\cite{liu2024improved} &  94.8 & 69.3 & 85.5 & 83.2 \\
C2P-CLIP~\cite{tan2025c2p}             & 57.4       & 64.2           & 54.2              & 58.6      \\ 
Qwen3-VL 2B$^\dagger$~\cite{Qwen3-VL}  & 88.9  & 89.9 & 68.8 & 82.5  \\
Qwen3-VL 4B$^\dagger$~\cite{Qwen3-VL}  & 96.0  & 89.5 & 71.3  & 85.6 \\
\hline
\multicolumn{5}{c}{\itbf{w/ our TranX-Adapter}} \\ \hline
Qwen3-VL 2B &     97.5         & 95.3   &  78.3  &  88.9   \\ 
\tablerowcolor \textbf{Qwen3-VL 4B} &     \textbf{98.1}         & \textbf{95.5}   & \textbf{79.0}  &  \textbf{90.9}   \\
LLaVA-1.6-mistral 7B &     96.6         &       93.0         &      77.1             &       88.9       \\ 
\bottomrule
\end{tabular}
}
\label{table:RRDataset}
\vspace{-0.4cm}
\end{table}

\subsection{Ablation Study}

To elucidate the contribution of each component within our \itbf{TranX-Adapter}, we perform ablation studies on GenImage using LLaVA-1.6-mistral 7B. As presented in Table~\ref{tab:ablation}, incorporating the artifact encoder to provide auxiliary forgery cues yields a 4.6\% improvement in accuracy. Introducing \textit{X-Fusion} and \textit{TOP-Fusion} individually brings further gains of 2.3\% and 3.2\%, respectively. When both fusion strategies are jointly applied, the model attains an accuracy of 91.9\%, confirming that the two fusion mechanisms complement each other and jointly enhance detection performance.

\begin{table}[htb]
\caption{\textbf{Ablation study of different module in \itbf{TranX-Adapter}}. The first row shows the results of LLaVA-1.6-mistral 7B. We integrate NPR, apply \textit{X-Fusion} and \textit{TOP-Fusion} separately, and then use the full bidirectional fusion strategy.}
\centering
\resizebox{1.\linewidth}{!}{
\begin{tabular}{cccc}
\toprule
\itbf{Artifact Encoder} & \itbf{X-Fusion} & \itbf{TOP-Fusion} & \itbf{Accuracy} \\ \hline
                 &          &            & 82.3     \\
        \checkmark         &          &            & 86.0     \\
          \checkmark       &     \checkmark     &            & 89.3     \\
        \checkmark         &          &    \checkmark        & 90.3     \\ \hdashline
 \tablerowcolor      \checkmark          &     \checkmark     &    \checkmark        & \textbf{91.9}    \\
\bottomrule
\end{tabular}
}
\label{tab:ablation}
\vspace{-0.4cm}
\end{table}

\subsection{Comparison with PEFT Methods}

To evaluate training efficiency, we compare our \itbf{TranX-Adapter} with Parameter-Efficient Fine-Tuning (PEFT) methods, including LoRA~\cite{hulora} and Adapter~\cite{houlsby2019parameter}, on the Chameleon dataset, using GenImage-SD v1.4 as the training set. As shown in Table~\ref{tab:compare_peft}, our \itbf{TranX-Adapter} achieves performance comparable to full fine-tuning while using only a tiny-scale of the parameters.

\begin{table}[htb]
\caption{\textbf{Comparison with PEFT methods on Chameleon dataset}, where ``Params.'' denotes the number of learnable parameters measured in millions.}
\centering
\begin{tabular}{lcc}
\toprule
\itbf{Method}                         & \itbf{Params. (M)} & \itbf{Acc.} \\ \hline
\multirow{1}{*}{Full}          &    7261       &  \textbf{76.8} \\ \hdashline

\multirow{2}{*}{LoRA~\cite{hulora}}          & 40        & 69.1     \\
                               & 160       & 74.4     \\ \hdashline
\multirow{2}{*}{Adapter~\cite{houlsby2019parameter}}       & 40        & 69.4     \\
                               & 160       & 72.5     \\ \hdashline
  \multirow{2}{*}{\itbf{TranX-Adapter}} & 40        & 73.8     \\
                      & 160       & 75.8 \\
\bottomrule
\end{tabular}
\label{tab:compare_peft}
\vspace{-0.4cm}
\end{table}

\begin{figure*}[thbp]
    \begin{center}
        \includegraphics[width=1.\linewidth]{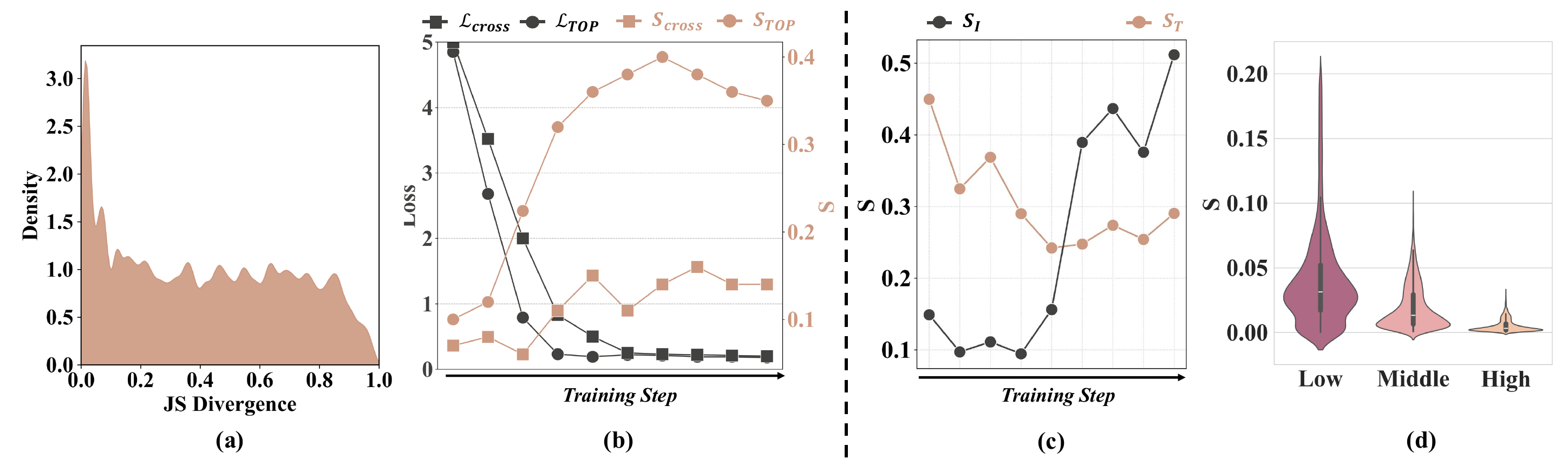}
    \end{center}
    \caption{\textbf{Overview of JS divergence and the evolution of information-flow significance across training.} (a) Distribution of JS divergence. (b) Illustrates the relationship between the training loss and the information-flow significance $S$. (c) Comparison of image and text information-flow significance and (d) layer-wise distribution of $S$ across the LLM, where ``Low'' indicates layers near the input.}
    \label{fig:discussion}
    \vspace{-0.2cm}
\end{figure*}
\subsection{Why does Our TranX-Adapter Work?}
\label{sec:discussion}
In Sect.\ref{sec:pilot_study}, we observe that the artifact features exhibit high intra-feature similarity, preventing effective injection of artifact information into the semantic features. In this section, we analyze why \itbf{TranX-Adapter} is effective from the following two perspectives:

\noindent\textbf{\textit{1) Why can TOP-Fusion transfer artifact information into semantic features more effectively?}}
We design \textit{TOP-Fusion} to selectively amplify regions where artifact and semantic diverge in AIGI detection. To achieve this, we compute the JS divergence between their prediction probabilities and adopt it as the cost matrix within the optimal transport. As illustrated in Figure~\ref{fig:discussion} (a), the JS divergence exhibits low JS divergence regions dominate with high density, whereas high JS divergence regions are relatively sparse. This non-uniformity easily guides the model to focus on patches with substantial discrepancies. To further assess the advantages of \textit{TOP-Fusion}, we compare it against a standard cross-attention by substituting \textit{TOP-Fusion} with cross-attention. We denote the mean significance of information flow~\cite{wang2023label} from artifact to semantic features as $S_{TOP}$ and $S_{cross}$ for \textit{TOP-Fusion} and cross-attention, respectively. As shown in Figure~\ref{fig:discussion} (b), $S_{TOP}$ consistently surpasses $S_{cross}$, and \textit{TOP-Fusion} achieves a lower training loss under identical training steps. These findings demonstrate that \itbf{TOP-Fusion more effectively highlights regions with substantial discrepancies between artifact and semantic features, thereby enabling more efficient fusion.}
\vspace{+0.1cm}

\noindent\textbf{\textit{2) Why can TranX-Adapter remain effective with only a small number of trainable parameters?}}
To answer this question, it is essential to clarify what the LLM primarily learns during training. We therefore examine how different features, including text embeddings, artifact features, and semantic features, interact within the LLM. A naive concatenation strategy is adopted, where all three feature types are provided jointly as input while fine-tuning the LLM. We denote $S_{T}$ and $S_{I}$ as the mean significance of information flow~\cite{wang2023label} from text embeddings and visual features to the final output, respectively. As shown in Figure~\ref{fig:discussion} (c), the model increasingly depends on visual features as training progresses, indicating that \itbf{the LLM places greater emphasis on visual feature interactions.} Furthermore, Figure~\ref{fig:discussion} (d) shows that \itbf{these visual feature interactions predominantly emerge in the shallow layers (corresponding to ``Low'') of the LLM}. This behavior aligns with the design of \itbf{TranX-Adapter}, which promotes effective visual feature interaction via \textit{TOP-Fusion} and \textit{X-Fusion} with minimal additional parameters.

\section{Conclusion}
In this paper, we introduce \itbf{TranX-Adapter}, a lightweight fusion adapter that effectively integrates semantic and artifact features to strengthen the AIGI detection capability of MLLMs. Extensive experiments across diverse benchmarks demonstrate that \itbf{TranX-Adapter} delivers robust performance and superior generalization. Our analysis reveals three key insights: (1) high intra-feature similarity of artifact features leads to \texttt{attention dilution} under self-attention, while discrepancy-aware fusion is more effective; (2) the LLM increasingly relies on visual information during training and (3) artifact-semantic fusion predominantly occurs in shallow layers.
Overall, \itbf{TranX-Adapter} improves artifact utilization in MLLMs, paving the way for future work on AIGI localization and explainability.

\nocite{langley00}

\bibliography{example_paper}
\bibliographystyle{icml2026}

\newpage


\end{document}